\documentclass{article}
\usepackage{spconf,amsmath,graphicx}
\usepackage{multirow}
\usepackage{amsmath}
\usepackage{amsfonts}

\newcommand{\ie}{\textit{i.e.}}

\newcommand{\etal}{\textit{et al.~}}

\title{Learning Regional Attention over Multi-resolution Deep Convolutional Features for Trademark Retrieval}
\name{Osman Tursun\sthanks{Corresponding Author. Mail: w.tuerxun@qut.edu.au}, Simon Denman, Sridha Sridharan, Clinton Fookes}
\address{Signal Processing, Artificial Intelligence and Vision Technologies (SAIVT)
\\ Queensland University of Technology, Australia}
\begin{document}
\maketitle
\begin{abstract}
Large-scale trademark retrieval is an important content-based image retrieval task. A recent study shows that off-the-shelf deep features aggregated with Regional-Maximum Activation of Convolutions (R-MAC) achieve state-of-the-art results. However, R-MAC suffers in the presence of background clutter/trivial regions and scale variance, and discards important spatial information. We introduce three simple but effective modifications to R-MAC to overcome these drawbacks. First, we propose the use of both sum and max pooling to minimise the loss of spatial information. We also employ domain-specific unsupervised soft-attention to eliminate background clutter and unimportant regions. Finally, we add multi-resolution inputs to enhance the scale-invariance of R-MAC. We evaluate these three modifications on the million-scale METU dataset. Our results show that all modifications bring non-trivial improvements, and surpass previous state-of-the-art results.
\end{abstract}

\section{introduction}
\label{sec:intro}
A trademark (logo) is one of the most valuable intellectual properties of a company or individual. All trademarks require registration to avoid reputational and profit damages caused by trademark infringements. A trademark will be registered only if no duplication is found when it is compared with other registered trademarks. However, the exponential increase in the total number of trademark registrations and applications has made the registration process challenging. According to statistics reported by the world intellectual property office (WIPO), 11.5 million trademark applications were filed worldwide in 2019, which is a 5.8\% increase over 2018 \cite{WIPO}.

Large-scale trademark retrieval (LSTR) systems have been developed to detect and prevent trademark infringements. Early LSTR systems are text or code-based systems (\ie Vienna System), where each trademark is captioned by a human expert. Later, LSTR using content-based image retrieval (CBIR) algorithms have been used thanks to it's efficiency and accuracy. Hand-crafted features based-on shape, color or texture were developed for early CBIR-LSTR systems \cite{tursun2015metu,tursun2019componet}. With the rise of deep learning, off-the-shelf deep features have been applied for LSTR, demonstrating higher accuracy and efficiency compared to traditional hand-crafted features.

Due to the lack of publicly available labeled trademark datasets, recent studies focus on improving LSTR with deep features via post-processing off-the-shelf deep features, or pre-processing inputs rather than fine-tuning a pre-trained network. This type of image retrieval study is defined as ``pre-trained single-pass" by Zheng \etal \cite{zheng2017sift}, and our work also belongs to this category. The recent state-of-the-art (SOTA) LSTR study \cite{tursun2019componet} shows R-MAC (Regional-Maximum Activation of Convolutions) \cite{tolias2015particular} is efficient, simple and accurate compared to other post-processing techniques \cite{babenko2015aggregating,kalantidis2016cross,Jimenez_2017_BMVC}. However, its results suffer when text-components appear in trademarks as they increase the number of regions with text or background that are not essential for similarity detection. Previous work  \cite{tursun2019componet} improved R-MAC results by removing text-components that appeared in trademarks, and this approach can be considered as hard-attention. Kim \etal \cite{kim2018regional} also claim that R-MAC suffers in the presence of background clutter and regions of varying importance. They improve R-MAC performance by applying regional context-aware soft-attention for each regional MAC feature. However, they generate this attention via a regional attention network that requires supervised training. Although they trained the regional attention network on ImageNet \cite{russakovsky2015imagenet}, the domain difference between the target domain and ImageNet should be taken into consideration.

In this work, we also improve R-MAC's performance using soft-attention, where soft-attention is learned in an unsupervised manner on the gallery images. We apply a bag-of-words (BoW) model \cite{sivic2003video} to temporal regional features in the R-MAC pipeline, so each region is viewed as a word, and each trademark as a document. Later, we calculate the inverse document frequency (IDF) \cite{salton1988term} value for each word, which is used as soft-attention for regional features.

Additionally, we replace the MAC pooling adopted in R-MAC with a concatenation of sum and max pooling. The max pooling used by R-MAC only selects the spatially maximum activations of convolutional features, that results in a loss of other spatial information. We therefore integrate sum and max pooling. This extra step only increases the dimension of temporal regional features, however, the dimension of the final feature after aggregation remains the same as the feature dimension is reduced with a post-processing operation such as $l_2$ normalisation and PCA-whitening \cite{jegou2012negative}. Finally, to combat scale difference in TR, we introduce multi-resolution \cite{seddati2017towards,gordo2017end} inputs to the R-MAC pipeline. 

We have tested our method on the challenging METU trademark dataset \cite{metudeeptursun}. All modifications show non-trivial improvements, and our final system surpasses the existing state-of-the-art methods by a large margin.

\section{Related Studies}
\label{sec:lit}
The most recent trademark retrieval works are based on off-the-shelf deep features, as deep features are efficient and more accurate than hand-crafted features \cite{metudeeptursun,aker2017analyzing}. However, due to the lack of an annotated trademark dataset, most approaches are ``pre-trained single pass" methods. Early works \cite{metudeeptursun, aker2017analyzing} deploy pre-trained deep features from fully-connected layers for trademark retrieval. Later, aggregated deep-features from convolutional layers are studied \cite{lan2017similar,tursun2019componet}. Lan \etal \cite{lan2017similar} apply uniform local binary patterns (LBP) as an aggregator. They showed improved results compared to deep-features from full-connected layers, although the aggregation method is slow and returns features with high dimensionality. In comparison, Tursun \etal \cite{tursun2019componet} tested popular aggregation methods including SPoC, MAC, CRoW and R-MAC on the METU dataset. Deep convolutional features with these aggregation methods not only achieved improved results, but also reduced the feature size. To enable further improvements, Tursun \etal \cite{tursun2019componet} proposed soft and hard attention methods.

Few studies that fine-tune deep networks exist in the literature. Perez \etal \cite{perez2018trademark} improved deep feature performance by fine-tuning deep features with classification loss. To achieve this, they built a visual similarity dataset with 151 classes and a conceptual similarity dataset with 205 classes. Lan \etal \cite{lan2018similar} also improved deep feature performance by fine-tuning deep features with the triplet loss. Xia \etal \cite{xia2019trademark} proposed a transformation-invariant deep hashing method for efficient and transform-invariant trademark retrieval. However, the training sets of these works are either private or a subset of the METU query set. This makes their results difficult to fairly compare with the results presented here.

\section{Multi-Scale Unsupervised Regional Attentive Deep Feature}
\label{sec:method}

\begin{figure*}[!ht]
	\centering
	\includegraphics[width=0.9\linewidth]{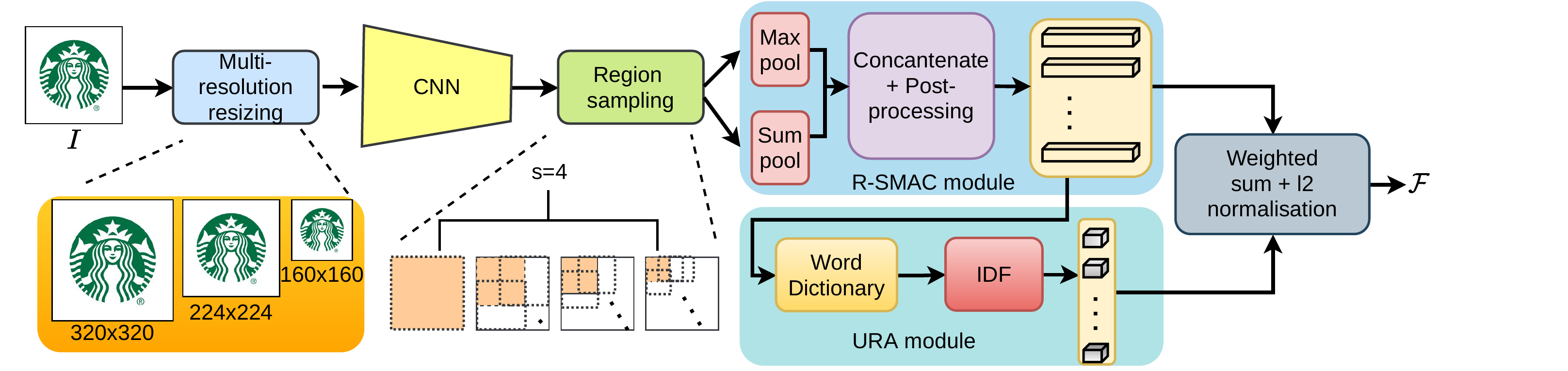}
	\caption{Overall diagram of the proposed method. An input image with three different resolutions ($160\times160$, $224\times224$ and $320\times320$) is sent to a Convolutional Neural Network (CNN) to extract features from a convolutional layer. In this work, all features are extracted from ResNet50/Conv4 \cite{he2016deep}. These features are processed with the R-SMAC module and unsupervised regional attention module (URA). Finally, features are aggregated through the weighted sum. Weight factors are calculated with the URA module.}
	\label{fig:ms-rmac}
\end{figure*}

In this section, we introduce three modifications to the R-MAC (Regional-Maximum Activation of Convolutions) pipeline to improve its performance for TR. They are multi-resolution (MR), sum and max activation of convolution (SMAC), and unsupervised regional attention (URA). As shown in Fig. \ref{fig:ms-rmac}, MR is placed in the front of the R-MAC pipeline, while SMAC and URA are at the end. Before presenting details of these modifications, we briefly introduce the R-MAC pipeline.

R-MAC is an advanced version of MAC that is widely used as an aggregator for convolutional features. Several CBIR studies show that mid-level convolutional feature maps with maximum or sum outperform fully-connected layer features \cite{tolias2015particular,Jimenez_2017_BMVC,kalantidis2016cross,babenko2015aggregating}, as they contain more general features. However, MAC sacrifices local spatial information for compactness. For example, a convolution feature map of Image $I$ is $\mathcal{X}$. The shape of $\mathcal{X}$ is  $C (channel) \times W  (width) \times H (height)$. MAC will perform a spatial maximum pooling for each channel $\mathcal{X}_c$ of $\mathcal{X}$. Therefore, the MAC of $\mathcal{X}$ is,

\begin{equation}
	\mathbf{f} = [f_1, \dots, f_c, \dots, f_C], with f_c =  \max_{x\in\mathcal{X}_c} x.
\end{equation}

The MAC operation generates a compact representation $\mathbf{f}$ of size $C$, but it discards all information except maximum values. To retain important local information, Tolias \etal \cite{tolias2015particular} sampled multi-scale square regions from $\mathcal{X}$ in a sliding window fashion as shown in the ``region sampling" module of Fig. \ref{fig:ms-rmac}. The width of the sliding window is $2\times min(W,H)/(s+1), s = 1\dots S$ and its stride ($S$)is 60\% of it's width. The total number of regions, $N$, is decided by $S$. In our experiment, $S$ is set to 4, $N$, therefore, is $30$. Here, we use the notation $\mathbf{f_{R_i}} = [f_{R_i, 1}, \dots, f_{R_i, c}, \dots, f_{R_i, C}]$ for the MAC feature of the region $i$.

The R-MAC feature is the sum aggregation of $N$ regional MAC features. Usually, prior to sum aggregation, post-processing such as $\l_2$-normalisation and PCA-Whitening \cite{jegou2012negative} is applied. Here, we use the notation `` $\hat{}$ '' to represent this post-processing. The R-MAC feature thus defined as,
\begin{equation}
	\mathcal{F} = \sum_{i=1}^{N}\hat{\mathbf{f}}_{R_i}.
\end{equation}

In experiments, we learned the PCA-whitening on the sampled 30,000 trademarks, and the feature size after PCA is set to 256.

{\bf{Sum and Max Activation of Convolution (SMAC)}} MAC only encodes the maximum ``local'' response of each of the convolutional filters, which causes the loss of other important information embedded in the convolution features. To overcome this, we also apply sum pooling over regional features in addition to max pooling, and use the region-wise concatenation of them. This doubles the dimension size of the temporal regional features, however after post-processing the dimension size is reduced to 256.

{\bf{Multi Resolution (MR)}} Recent works \cite{gordo2017end,seddati2017towards} have improved R-MAC performance by using multiple resolutions. Inspired by these, we extract three convolutional feature maps for an input image $I$ with three resolutions ($160\times160$, $224\times224$ and $320\times320$). We therefore obtain $3\times N$ temporal regional features. The final R-MAC is the sum aggregation of these. Note all temporal regional features are post-processed. With this, $\mathcal{F}$ is equal to,

\begin{equation}
	\mathcal{F} = \sum_{j=1}^{3} \sum_{i=1}^{N}\hat{\mathbf{f}}_{R_{i,j}},
\end{equation}

where $\hat{\mathbf{f}}_{R_{i,j}}$ represents $i$th temporal regional feature of $j$th resolution.

{\bf{Unsupervised Regional Attention (URA)}} Kim \etal \cite{kim2018regional} note that R-MAC suffers from background clutter and varying importance of regions. They applied context-aware soft-attention to overcome this drawback of R-MAC. However, they generate the context-aware soft-attention signal via a regional attention network that is trained with ImageNet \cite{russakovsky2015imagenet}, and their method is intended for tasks where no labelled dataset is available for the target domain. However, they don't consider the context-difference between ImageNet and the target domain. For example, what is considered to be an important region in ImageNet might be unimportant for the target domain say TR. Therefore, we propose an unsupervised regional attention method to generate domain-specific context-aware soft attention. Our method learns the domain-specific context-aware soft-attention in an unsupervised manner from the gallery images that are available for every image retrieval task.

To model this domain-specific context-awareness, we applied a bag-of-words (BoW) model \cite{sivic2003video}. We build a regional deep feature dictionary by clustering sampled regional deep features extracted from the gallery images. K-means clustering (using the  FAISS library \cite{JDH17}) is applied to build the dictionary of 1,024 words. Noting the wide use of term-frequency (TF) and inverse-document-frequency (IDF) statistics \cite{salton1988term} for text-retrieval and context modelling, we calculate TF-IDF values for each word here to obtain a form of soft-attention. However, we note that TF is implicitly utilised during the sum aggregation step in R-MAC. Therefore, we only calculated the IDF for each regional deep feature,
\begin{equation}
	\mathcal{F} = \sum_{j=1}^{3}\sum_{i=1}^{N}\hat{\mathbf{f}}_{R_{i,j}}*IDF(\hat{\mathbf{f}}_{R_{i,j}}).
\end{equation}

In our experiments, we measure the similarity of two trademarks by calculating the Euclidean distance between their $l_2$ normalised modified R-MAC features.

\section{Experiments}
\label{sec:exp}

\begin{figure*}[th!]
	\centering
	\includegraphics[width=0.9\linewidth]{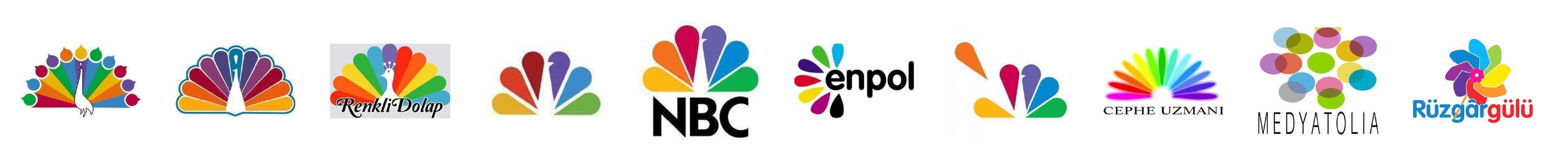}
	\includegraphics[width=0.9\linewidth]{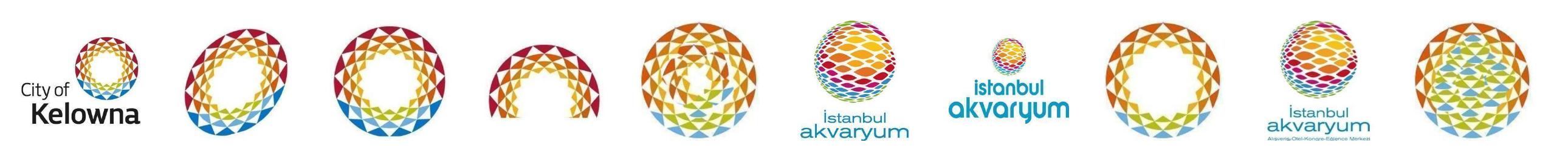}
	\includegraphics[width=0.9\linewidth]{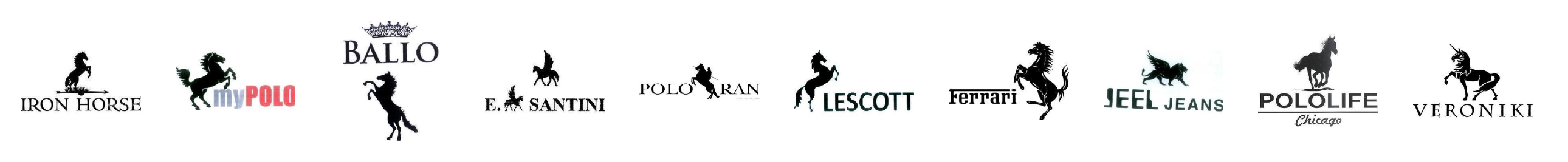}
	\caption{Top10 retrieved results from the METU dataset for example queries shown in the leftmost column.} 
	\label{fig:qual}
\end{figure*}

\subsection{Dataset and Evaluation Protocol}
We select the METU trademark retrieval dataset \cite{metudeeptursun} as our testing dataset. It is the largest public dataset for TR. It includes nearly 1 million trademarks composed of text-only marks, figure-only marks and figure and text marks. Its evaluation set is composed of 35 similar groups, and each group includes around 10 to 14 similar trademarks. In total it includes 417 queries.

For evaluation, we follow the same evaluation protocol described in \cite{tursun2015metu}. In detail, we first return the ranking results for each query by sorting the similarity scores of the gallery images. We evaluated performance using the normalized average rank (NAR) and mean average precision (MAP) metrics. The NAR is calculated by normalizing the average ranking position of the ground-truth results of the queries. MAP values are only calculated for the top 100 results.

\subsection{Comparison with State-of-the-Art Methods}
We compare our multi-resolution R-MAC with unsupervised regional attention method with the recent SOTA TR methods that have been tested on the METU trademark dataset. Comparison results are shown in Table \ref{tab:sota}. Systems are categorized into three groups: hand-crafted features, fine-tuned off-the-shelf deep features, and pre-trained single pass \cite{zheng2017sift,kim2018regional}. Our method belongs to the last group as we don't fine-tune the feature extraction network. Our method achieves state-of-the-art NAR and MAP@100 results. In addition, our methods feature dimension is 256, which is the same as the previous SOTA.

\begin{table}[h!]
	\small
	\begin{center}
		\begin{tabular}{l c  c  c}
			\hline
			\bf Method & \bf DIM $\downarrow$ & \bf NAR $\downarrow$ & \bf MAP@100 $\uparrow$\\ \hline
			\multicolumn{4}{c}{\bf hand-crafted features} \\\hline
			Feng \etal \cite{feng2018aggregation} & 6,224 &  0.083 &  - \\
			Tursun \etal \cite{metudeeptursun}   & 10k   &  0.062 &  - \\\hline
			\multicolumn{4}{c}{\bf fine-tuned off-the-shelf deep features} \\\hline
			Perez  \etal (vis) \cite{perez2018trademark} & 4,096  & 0.066  & - \\
			Perez  \etal (con) \cite{perez2018trademark} & 4,096 & 0.063 & - \\
			Perez  \etal (vis, con) \cite{perez2018trademark} & 4,096  & 0.047 & - \\\hline
			\multicolumn{4}{c}{\bf pre-trained single pass \cite{zheng2017sift,kim2018regional}} \\\hline
			SPoC \cite{babenko2015aggregating,tursun2019componet} &  256 & 0.120 & 18.7\\ 
			CRoW \cite{kalantidis2016cross,tursun2019componet} 			& 256   &  0.140 & 19.8     \\
			R-MAC \cite{tolias2015particular}  		& 256 	& 0.072 & 24.8 \\
			MAC \cite{tolias2015particular,tursun2019componet} 	& 512  & 0.120 & 21.5 \\
			Jimenez \cite{Jimenez_2017_BMVC,tursun2019componet}   & 256 	& 0.093 & 21.0  \\
			CAM MAC \cite{tursun2019componet}  &  256  & 0.064   &22.3  \\
			ATR MAC \cite{tursun2019componet} &  512  & 0.056   &24.9 \\
			ATR R-MAC \cite{tursun2019componet} &	256  & 0.063	& 25.7 \\
			ATR CAM MAC \cite{tursun2019componet}	&  512  & 0.040   &  25.1 \\
			MR-R-MAC w/UAR (ours)  &  256  &  \textbf{0.028}  & \textbf{30.6} \\\hline
		\end{tabular}
	\end{center}
	\caption{Comparison with the previous state-of-the-art results on the METU dataset. NAR is the normalized average rank metric.}
	\label{tab:sota}
\end{table}

\begin{table}[h!]
	\small
	\begin{center}
		\begin{tabular}{l | c  c c }
			\hline
			\bf Method & \bf DIM  $\downarrow$ & \bf NAR $\downarrow$ & \bf MAP@100 $\uparrow$ \\ \hline
			R-MAC \cite{tursun2019componet}  &  256  & 0.051 & 27.3 \\
			R-SMAC \cite{tursun2019componet}  &  256 & 0.045 & 27.8 \\
			MR-R-MAC  \cite{tursun2019componet} &256 & 0.040 & 29.1 \\
			R-MAC w/URA & 256   & 0.039 & 29.6 \\
			MR-R-MAC w/URA & 256  & 0.033 &  30.6  \\
			MR-R-SMAC w/URA & 256  & \bf 0.028 &  \bf 31.0  \\
			\hline
		\end{tabular}
	\end{center}
	\label{tab:abl}
	\caption{Ablation study for the proposed approach.}
\end{table}

\subsection{Ablation Study}
We conduct an ablation study to consider the three modifications introduced to the R-MAC pipeline: multi-resolution (MR), sum and max pooling (R-SMAC), and unsupervised attention (UAR). We evaluate all possible combinations of the three modifications. Results are given in Table \ref{tab:abl}. All modifications bring non-trivial improvements. Moreover, the results show that they are complementary to one another. MR-R-SMAC with UAR (all three modifications) achieves the best result.

\subsection{Qualitative Results}
In Fig. \ref{fig:qual}, we visualized the top 10 retrieved results for three sampled queries. Here, all of the retrieved results are highly similar or related to the corresponding query. For the figure-only query displayed in the top row, figure-text marks are listed in the top 10 queries. On the other hand, for figure-text queries in the middle and bottom rows, no text-only marks are returned in the top 10 results. We therefore conclude that the proposed domain-specific context-aware unsupervised attention is beneficial.
\section{Conclusion}
\label{sec:con}
We have introduced three modifications for the R-MAC pipeline to improve trademark retrieval performance: the use of multi-resolution input; domain-specific context-aware regional attention; and the use of both sum and max pooling for feature aggregation. With them, R-MAC features are more discriminant and robust to background-clutter and changes in scale. In the METU trademark dataset, the proposed method achieves a $5.3\%$ improvement over the state-of-the-art in MAP@100 metric and decreases the NAR score by 0.012.

\bibliographystyle{IEEEbib}
\bibliography{egbib}

\end{document}